\documentclass[11pt,a4paper]{article}
\usepackage{acl2018}
\usepackage{times}
\usepackage{latexsym}

\usepackage{bbm}
\usepackage{multirow}
\usepackage{booktabs}
\usepackage{url}
\usepackage{xcolor}
\usepackage{graphicx}
\usepackage{tabularx}
\usepackage{verbatim}
\usepackage{enumitem,kantlipsum}

\usepackage{amsmath}
\usepackage{amssymb}

\usepackage{bm}
\newcommand{\M}[2]{\bm{#1}_{#2}}
\newcommand{\V}[1]{\bm{#1}}

\aclfinalcopy 

\title{
Long Short-Term Memory\\as a Dynamically Computed Element-wise Weighted Sum
}

\author{
Omer Levy\thanks{The first two authors contributed equally to this paper.}
 \qquad Kenton Lee$^*$
 \qquad Nicholas FitzGerald
 \qquad Luke Zettlemoyer\\
Paul G. Allen School, University of Washington, Seattle, WA\\
\texttt{\{omerlevy,kentonl,nfitz,lsz\}@cs.washington.edu}
}

\begin{document}

\maketitle

\begin{abstract}
LSTMs were introduced to combat vanishing gradients in simple RNNs by augmenting them with gated additive recurrent connections.
We present an alternative view to explain the success of LSTMs: 
the gates themselves are versatile recurrent models that provide more representational power than previously appreciated. 
We do this by decoupling the LSTM's gates from the embedded simple RNN, producing a new class of RNNs where the recurrence computes an element-wise weighted sum of context-independent functions of the input. 
Ablations on a range of problems demonstrate that the gating mechanism alone performs as well as an LSTM in most settings, strongly suggesting that the gates are doing much more in practice than just alleviating vanishing gradients.
\end{abstract}

\section{Introduction}

Long short-term memory (LSTM)~\cite{lstm} has become the de-facto recurrent neural network (RNN) for learning representations of sequences in NLP.  
Like simple recurrent neural networks (S-RNNs)~\cite{rnn}, LSTMs are able to learn non-linear functions of arbitrary-length input sequences. However, they also introduce an additional memory cell to mitigate the vanishing gradient problem \cite{hochreiter:91,vangrad}. This memory is controlled by a mechanism of 
gates, whose additive connections allow long-distance dependencies to be learned  more easily
during backpropagation. While this view is mathematically accurate, in this paper 
we argue that it does not provide a complete picture of why LSTMs work in practice.

We present an alternate view to explain the success of LSTMs: the gates themselves are powerful recurrent models that provide more representational power than previously realized. 
To demonstrate this, we first show that LSTMs can be seen as a combination of two recurrent models: (1) an S-RNN, and (2) an element-wise weighted sum of the S-RNN's outputs over time, which is implicitly computed by the gates.
We hypothesize that, for many practical NLP problems, the weighted sum serves as the main modeling component. The S-RNN, while theoretically expressive, is in practice only a minor contributor that clouds the mathematical clarity of the model. By replacing the S-RNN with a context-\emph{independent} function of the input, we arrive at a much more restricted class of RNNs, where the main recurrence is via the element-wise weighted sums that the gates are computing.

We test our hypothesis on NLP problems, where LSTMs are wildly popular at least in part due to their ability to model crucial phenomena such as word order~\cite{adi2017}, syntactic structure~\cite{linzen2016}, and even long-range semantic dependencies~\cite{luhengsrl}.
We consider four challenging tasks: language modeling, question answering, dependency parsing, and machine translation. 
Experiments show that while removing the gates from an LSTM can severely hurt performance, replacing the S-RNN with a simple linear transformation of the input results in minimal or no loss in model performance.
We also show that, in many cases, LSTMs can be further simplified by removing the output gate, arriving at an even more transparent architecture, where the output is a context-\emph{independent} function of the weighted sum.
Together, these results suggest that the gates' ability to compute an element-wise weighted sum, rather than the non-linear transition dynamics of S-RNNs, are the driving force behind LSTM's success.

\section{What Do Memory Cells Compute?}
\label{sec:theoretical}

LSTMs are typically motivated as an augmentation of simple RNNs (S-RNNs), defined as:
\begin{align}
    \V{h}_{t}&=\tanh(\M{W}{hh}\V{h}_{t-1} + \M{W}{hx}\V{x}_t  + \V{b}_{h})
\end{align}
S-RNNs suffer from the vanishing gradient problem~\cite{hochreiter:91,vangrad} due to compounding multiplicative updates of the hidden state. By introducing a memory cell and an output layer controlled by gates, LSTMs enable shortcuts through which gradients can flow when learning with backpropagation. This mechanism enables learning of long-distance dependencies while preserving the expressive power of recurrent non-linear transformations provided by S-RNNs.

Rather than viewing the gates as simply an auxiliary mechanism to address a \emph{learning} problem, we present an alternate view that emphasizes their \emph{modeling} strengths. We argue that the LSTM should be interpreted as a hybrid of two distinct recurrent architectures: (1) the S-RNN which provides multiplicative connections across timesteps, and (2) the memory cell which provides additive connections across timesteps. On top of these recurrences, an output layer is included that simply squashes and filters the memory cell at each step.

Throughout this paper, let $\{\V{x}_1, \ldots, \V{x}_n\}$ be the sequence of input vectors, $\{\V{h}_1, \ldots, \V{h}_n\}$ be the sequence of output vectors, and $\{\V{c}_1, \ldots, \V{c}_n\}$ be the memory cell's states. Then, given the basic LSTM definition below, we can formally identify three sub-components. 
\begin{align}
    \widetilde{\V{c}}_{t}&=\tanh(\M{W}{ch}\V{h}_{t-1} + \M{W}{cx}\V{x}_t  + \V{b}_{c})\label{eq:content_layer}\\
    \V{i}_{t}&=\sigma(\M{W}{ih}\V{h}_{t-1} + \M{W}{ix}\V{x}_t + \V{b}_{i})\label{eq:input_gates}\\
    \V{f}_{t}&=\sigma(\M{W}{fh}\V{h}_{t-1} + \M{W}{fx}\V{x}_t + \V{b}_{f})\\
    \V{c}_{t}&=\V{i}_{t}\circ \widetilde{\V{c}}_{t}+\V{f}_{t} \circ \V{c}_{t-1} \label{eq:cell_update}\\
    \V{o}_{t}&=\sigma(\M{W}{oh} \V{h}_{t-1} + \M{W}{ox}\V{x}_t + \V{b}_{o})\label{eq:output_gates}\\
    \V{h}_{t}&=\V{o}_{t}\circ \tanh(\V{c}_{t})\label{eq:output_layer}
\end{align}
\paragraph{Content Layer (Equation~\ref{eq:content_layer})}
We refer to $\widetilde{\V{c}}_{t}$ as the content layer, which is the output of an S-RNN. Evaluating the need for multiplicative recurrent connections in the content layer is the focus of this work. The content layer is passed to the memory cell, which decides which parts of it to store.

\paragraph{Memory Cell (Equations~\ref{eq:input_gates}-\ref{eq:cell_update})}
The memory cell $\V{c}_{t}$ is controlled by two gates. The input gate $\V{i}_t$ controls what part of the content ($\widetilde{\V{c}}_{t}$) is written to the memory, while the forget gate $\V{f}_t$ controls what part of the memory is deleted by filtering the previous state of the memory ($\V{c}_{t-1}$). Writing to the memory is done by adding the filtered content ($\V{i}_{t}\circ \widetilde{\V{c}}_{t}$) to the retained memory ($\V{f}_{t} \circ \V{c}_{t-1}$).

\paragraph{Output Layer (Equations~\ref{eq:output_gates}-\ref{eq:output_layer})}
The output layer $\V{h}_{t}$ passes the memory cell through a $\tanh$ activation function and uses an output gate $\V{o}_t$ to read selectively from the squashed memory cell.

Our goal is to study how much each of these components contribute to the empirical performance of LSTMs. In particular, it is worth considering the memory cell in more detail to reveal why it could serve as a standalone powerful model of long-distance context. It is possible to show that it implicitly computes an \emph{element-wise weighted sum} of all the previous content layers by expanding the recurrence relation in Equation \ref{eq:cell_update}:
\begin{align}
\begin{split}
\V{c}_{t}&=\V{i}_{t}\circ \widetilde{\V{c}}_{t}+\V{f}_{t} \circ \V{c}_{t-1}\\
&= \sum_{j=0}^{t}\Big(\V{i}_{j}\circ \prod_{k=j+1}^{t} \V{f}_k \Big) \circ \widetilde{\V{c}}_{j}\\
&= \sum_{j=0}^{t}\V{w}_{j}^{t} \circ \widetilde{\V{c}}_{j}
\label{eq:weighted_sum}
\end{split}
\end{align}
Each weight $\V{w}_{j}^{t}$ is a product of the input gate $\V{i}_{j}$ (when its respective input $\widetilde{\V{c}}_{j}$ was read) and every subsequent forget gate $\V{f}_{k}$. An interesting property of these weights is that, like the gates, they are also soft element-wise binary filters.

\section{Standalone Memory Cells are Powerful}

The restricted space of element-wise weighted sums allows for easier mathematical analysis, visualization, and perhaps even learnability. However, constrained function spaces are also less expressive, and a natural question is whether these models will work well for NLP problems that involve understanding context. We hypothesize that the memory cell (which computes weighted sums) can function as a standalone contextualizer. To test this hypothesis, we present several simplifications of the LSTM's architecture (Section~\ref{subsec:models}), and show on a variety of NLP benchmarks that there is a qualitative performance difference between models that contain a memory cell and those that do not (Section~\ref{subsec:results}). We conclude that the content and output layers are relatively minor contributors, and that the space of element-wise weighted sums is sufficiently powerful to compete with fully parameterized LSTMs (Section~\ref{subsec:discussion}).

\subsection{Simplified Models}
\label{subsec:models}


The modeling power of LSTMs is commonly assumed to derive from the S-RNN in the content layer, with the rest of the model acting as a learning aid to bypass the vanishing gradient problem. We first isolate the S-RNN by ablating the gates (denoted as \emph{LSTM -- GATES} for consistency).

To test whether the memory cell has enough modeling power of its own, we take an LSTM and replace the S-RNN in the content layer from Equation~\ref{eq:content_layer} with a simple linear transformation ($\widetilde{\V{c}}_{t} = \M{W}{cx}\V{x}_t$) creating the \emph{LSTM -- S-RNN} model.

We further simplify the LSTM by removing the output gate from Equation~\ref{eq:output_layer} (${h}_{t} = \tanh(\V{c}_{t})$), leaving only the activation function in the output layer (\emph{LSTM -- S-RNN -- OUT}).
After removing the S-RNN and the output gate from the LSTM, the entire ablated model can be written in a modular, compact form:
\begin{align}
\begin{split}
\V{h}_{t}&= \textsc{output}\Big(\sum_{j=0}^{t}\V{w}_{j}^{t} \circ \textsc{content}(\V{x}_{j})\Big)
\label{eq:weighted_sum_property}
\end{split}
\end{align}
where the content layer $\textsc{content}(\cdot)$ and the output layer $\textsc{output}(\cdot)$ are both context-\emph{independent} functions, making the entire model highly constrained and mathematically simpler. The complexity of modeling contextual information is needed only for computing the weights $\V{w}_{j}^{t}$. As we will see in Section~\ref{subsec:results}, both of these ablations perform on par with LSTMs on several tasks. 

Finally, we ablate the hidden state from the gates as well, by computing each gate $\V{g}_{t}$ via $\sigma(\M{W}{gx}\V{x}_t + \V{b}_{g})$. In this model, the only recurrence is the additive connection in the memory cell; it has no multiplicative recurrent connections at all. It can be seen as a type of QRNN \cite{qrnn} or SRU \cite{sru}, but for consistency we label it as \emph{LSTM -- S-RNN -- HIDDEN}.


\subsection{Experiments}
\label{subsec:results}

We compare model performance on four NLP tasks, with an experimental setup that is lenient towards LSTMs and harsh towards its simplifications. In each case, we use existing implementations and previously reported hyperparameter settings. Since these settings were tuned for LSTMs, any simplification that performs equally to (or better than) LSTMs under these LSTM-friendly settings provides strong evidence that the ablated component is not a contributing factor. For each task 
we also report the mean and standard deviation of 5 runs of the LSTM settings to demonstrate the typical variance observed due to training with different random initializations.

\paragraph{Language Modeling}
We evaluate the models on the Penn Treebank (PTB)~\cite{ptb} language modeling benchmark. We use the implementation of \newcite{zaremba2014} from TensorFlow's tutorial while replacing any invocation of LSTMs with simpler models. We test two of their configurations: \textit{medium} and \textit{large} (Table~\ref{tab:lm_results}).

\paragraph{Question Answering}
For question answering, we use two different QA systems on the Stanford question answering dataset (SQuAD)~\cite{squad}: the Bidirectional Attention Flow model (BiDAF)~\cite{bidaf} and DrQA~\cite{drqa}. BiDAF contains 3 LSTMs, which are referred to as the phrase layer, the modeling layer, and the span end encoder. Our experiments replace each of these LSTMs with their simplified counterparts. We directly use the implementation of BiDAF from AllenNLP~\cite{allennlp}, and all experiments reuse the existing hyperparameters that were tuned for LSTMs. Likewise, we use an open-source implementation of DrQA\footnote{\url{https://github.com/hitvoice/DrQA}} and replace only the LSTMs, while leaving everything else intact.
Table~\ref{tab:squad_results} shows the results.


\paragraph{Dependency Parsing}
For dependency parsing, we use the Deep Biaffine Dependency Parser~\cite{dozat:16}, which relies on stacked bidirectional LSTMs to learn context-sensitive word embeddings for determining arcs between a pair of words. We directly use their released implementation, which is evaluated on the Universal Dependencies English Web Treebank v1.3~\cite{ewt}. In our experiments, we use the existing hyperparameters and only replace the LSTMs with the simplified architectures.
Table~\ref{tab:parsing_results} shows the results.


\paragraph{Machine Translation}
For machine translation, we used OpenNMT \cite{opennmt} to train English to German translation models on the multi-modal benchmarks from WMT 2016 (used in OpenNMT's readme file). We use OpenNMT's default model and hyperparameters, replacing the stacked bidirectional LSTM encoder with the simplified architectures.\footnote{For the S-RNN baseline (\emph{LSTM -- GATES}), we had to tune the learning rate to 0.1 because the default value (1.0) resulted in exploding gradients. This is the only case where hyperparameters were modified in all of our experiments.}
Table~\ref{tab:translation_results} shows the results.

\begin{table}[t]
    \centering
    \small{
    \begin{tabular}{l l l}
        \toprule
        \textbf{Configuration} & \textbf{Model} & \textbf{Perplexity} \\
        \midrule
        \multirow{5}{*}{\shortstack[l]{PTB\\(Medium)}} 
        & LSTM & ~~83.9 $\pm$ 0.3 \\
        & -- S-RNN & ~~80.5 \\
        & -- S-RNN -- OUT & ~~81.6 \\
        & -- S-RNN -- HIDDEN & ~~83.3 \\
        & -- GATES & 140.9 \\
        \midrule
        \multirow{5}{*}{\shortstack[l]{PTB\\(Large)}} 
        & LSTM & ~~78.8 $\pm$ 0.2 \\
        & -- S-RNN & ~~76.0 \\
        & -- S-RNN -- OUT & ~~78.5 \\
        & -- S-RNN -- HIDDEN & ~~82.9 \\
        & -- GATES & 126.1 \\
        \bottomrule
    \end{tabular}
    }
    \caption{Performance on language modeling benchmarks, measured by perplexity.}
    \label{tab:lm_results}
\end{table}

\begin{table}[t]
    \centering
    \small{
    \begin{tabular}{l l l l}
        \toprule
        \textbf{System} & \textbf{Model} & \textbf{EM} & \textbf{F1} \\
        \midrule
        \multirow{4}{*}{BiDAF} & LSTM & 67.9 $\pm$ 0.3 & 77.5 $\pm$ 0.2 \\
        & -- S-RNN & 68.4 & 78.2 \\
        & -- S-RNN -- OUT & 67.4 & 77.2 \\
        & -- S-RNN -- HIDDEN & 66.5 & 76.6 \\
        & -- GATES & 62.9 & 73.3 \\
        \midrule
        \multirow{4}{*}{DrQA} & LSTM & 68.8 $\pm$ 0.2 & 78.2 $\pm$ 0.2 \\
        & -- S-RNN & 68.0 & 77.2 \\
        & -- S-RNN -- OUT & 68.7 & 77.9 \\
        & -- S-RNN -- HIDDEN & 67.9 & 77.2 \\
        & -- GATES & 56.4 & 66.5 \\
        \bottomrule
    \end{tabular}
    }
    \caption{Performance on SQuAD, measured by exact match (EM) and span overlap (F1).}
    \label{tab:squad_results}
\end{table}

\begin{table}[t]
    \centering
    \small{
    \begin{tabular}{l l l}
        \toprule
        \textbf{Model} & \textbf{UAS} & \textbf{LAS} \\
        \midrule
        LSTM & 90.60 $\pm$ 0.21 & 88.05 $\pm$ 0.33 \\
        -- S-RNN & 90.77 & 88.49\\
        -- S-RNN -- OUT & 90.70 & 88.31 \\
        -- S-RNN -- HIDDEN & 90.53 & 87.96 \\
        -- GATES & 87.75 & 84.61 \\
        \bottomrule
    \end{tabular}
    }
    \caption{Performance on the universal dependencies parsing benchmark, measured by unlabeled (UAS) and labeled attachment score (LAS).}
    \label{tab:parsing_results}
\end{table}

\begin{table}[t]
    \centering
    \small{
    \begin{tabular}{l l}
        \toprule
        \textbf{Model} & \textbf{BLEU} \\
        \midrule
        LSTM & 38.19 $\pm$ 0.1 \\
        -- S-RNN & 37.84 \\
        -- S-RNN -- OUT & 38.36 \\
        -- S-RNN -- HIDDEN & 36.98 \\
        -- GATES & 26.52 \\
        \bottomrule
    \end{tabular}
    }
    \caption{Performance on the WMT 2016 multi-modal English to German benchmark.}
    \label{tab:translation_results}
\end{table}

\subsection{Discussion}
\label{subsec:discussion}

We showed four major ablations of the LSTM.
In the S-RNN experiments (\emph{LSTM -- GATES}), we ablate the memory cell and the output layer. In the \emph{LSTM -- S-RNN} and \emph{LSTM -- S-RNN -- OUT} experiments, we ablate the S-RNN. In the \emph{LSTM -- S-RNN -- HIDDEN}, we remove not only the S-RNN in the content layer, but also the S-RNNs in the gates, resulting in a model whose sole recurrence is in the memory cell's additive connection.

As consistent with previous literature, removing the memory cell degrades performance drastically. In contrast, removing the S-RNN makes little to no difference in the final performance, suggesting that the memory cell alone is largely responsible for the success of LSTMs in NLP. 

Even after removing every multiplicative recurrence from the memory cell itself, the model's performance remains well above the vanilla S-RNN's, and falls within the standard deviation of an LSTM's on some tasks (see Table \ref{tab:parsing_results}).
This latter result indicates that the additive recurrent connection in the memory cell -- and not the multiplicative recurrent connections in the content layer or in the gates -- is the most important computational element in an LSTM. As a corollary, this result also suggests that a weighted sum of context words, while mathematically simple, is a powerful model of contextual information.

\section{LSTM as Self-Attention}

Attention mechanisms are widely used in the NLP literature to aggregate over a sequence \cite{cho2014,attention} or contextualize tokens within a sequence \cite{cheng2016,parikh2016} by \emph{explicitly} computing weighted sums. In the previous sections, we demonstrated that LSTMs implicitly compute weighted sums as well, and that this computation is central to their success. How, then, are these two computations related, and in what ways do they differ?

After simplifying the content layer and removing the output gate (\emph{LSTM -- S-RNN -- OUT}), the model's computation can be expressed as a weighted sum of context-independent functions of the inputs (Equation~\ref{eq:weighted_sum_property} in Section~\ref{subsec:models}). This formula abstracts over both the simplified LSTM and the family of attention mechanisms, and through this lens, the memory cell's computation can be seen as a ``cousin'' of self-attention. In fact, we can also leverage this abstraction to visualize the simplified LSTM's weights as is commonly done with attention (see Appendix~\ref{sec:visualization} for visualization).

However, there are three major differences in how the \emph{weights} $w_j^t$ are computed. 

First, the LSTM's weights are \emph{vectors}, while attention typically computes scalar weights; i.e. a separate weighted sum is computed for every dimension of the LSTM's memory cell. Multi-headed self-attention~\cite{vaswani2017} can be seen as a middle ground between the two approaches, allocating a scalar weight for different subsets of the dimensions.

Second, the weighted sum is accumulated with a dynamic program. This enables a linear rather than quadratic complexity in comparison to self-attention, but reduces the amount of parallel computation. This accumulation also creates an inductive bias of attending to nearby words, since the weights can only decrease over time.

Finally, attention has a probabilistic interpretation due to the softmax normalization, while the sum of weights in LSTMs can grow up to the sequence length. In variants of the LSTM that tie the input and forget gate, such as coupled-gate LSTMs \cite{greff2016} and GRUs \cite{cho2014}, the memory cell instead computes a weighted \emph{average} with a probabilistic interpretation. These variants compute locally normalized distributions via a product of sigmoids rather than globally normalized distributions via a single softmax.

\section{Related Work}

Many variants of LSTMs~\cite{lstm} have been previously explored. These typically consist of a different parameterization of the gates, such as LSTMs with peephole connections~\cite{gers2000}, or a rewiring of the connections, such as GRUs~\cite{cho2014}. However, these modifications invariably maintain the recurrent content layer. Even more systematic explorations~\cite{jozefowicz2015, greff2016,  zoph2017} do not question the importance of the embedded S-RNN. This is the first study to provide apples-to-apples comparisons between LSTMs with and without the recurrent content layer.

Several other recent works have also reported promising results with recurrent models that are vastly simpler than LSTMs, such as quasi-recurrent neural networks~\cite{qrnn}, strongly-typed recurrent neural networks~\cite{balduzzi:16}, recurrent additive networks~\cite{rans}, kernel neural networks~\cite{lei2017}, and simple recurrent units~\cite{sru}, making it increasingly apparent that LSTMs are over-parameterized. While these works indicate an obvious trend, they do not focus on explaining what LSTMs are learning. In our carefully controlled ablation studies, we propose and evaluate the minimal changes required to test our hypothesis that LSTMs are powerful because they dynamically compute element-wise weighted sums of content layers.

\section{Conclusion}

We presented an alternate view of LSTMs: they are a hybrid of S-RNNs and a gated model that dynamically computes weighted sums of the S-RNN outputs. Our experiments investigated whether the S-RNN is a necessary component of LSTMs. In other words, are the gates alone as powerful of a model as an LSTM? Results across four major NLP tasks (language modeling, question answering, dependency parsing, and machine translation) indicate that LSTMs suffer little to no performance loss when removing the S-RNN. This provides evidence that the gating mechanism is doing the heavy lifting in modeling context. We further ablate the recurrence in each gate and find that this incurs only a modest drop in performance, indicating that the real modeling power of LSTMs stems from their ability to compute element-wise weighted sums of context-independent functions of their inputs.

This realization allows us to mathematically relate LSTMs and other gated RNNs to attention-based models. Casting an LSTM as a dynamically-computed attention mechanism enables the visualization of how context is used at every timestep, shedding light on the inner workings of the relatively opaque LSTM.

\section*{Acknowledgements}
The research was supported in part by DARPA under the DEFT program (FA8750-13-2-0019), the ARO (W911NF-16-1-0121), the NSF (IIS-1252835, IIS-1562364), gifts from Google, Tencent, and Nvidia, and an Allen Distinguished Investigator Award. We also thank Yoav Goldberg, Benjamin Heinzerling, Tao Lei, and the UW NLP group for helpful conversations and comments on the work.

\bibliographystyle{acl_natbib}
\bibliography{references}

\appendix

\section{Weight Visualization}
\label{sec:visualization}

\begin{table*}[t]
    \centering
    \begin{tabular}{c c}
        \toprule
        Language model weights & Dependency parser weights \\
        \midrule
    \includegraphics[width=0.46\linewidth]{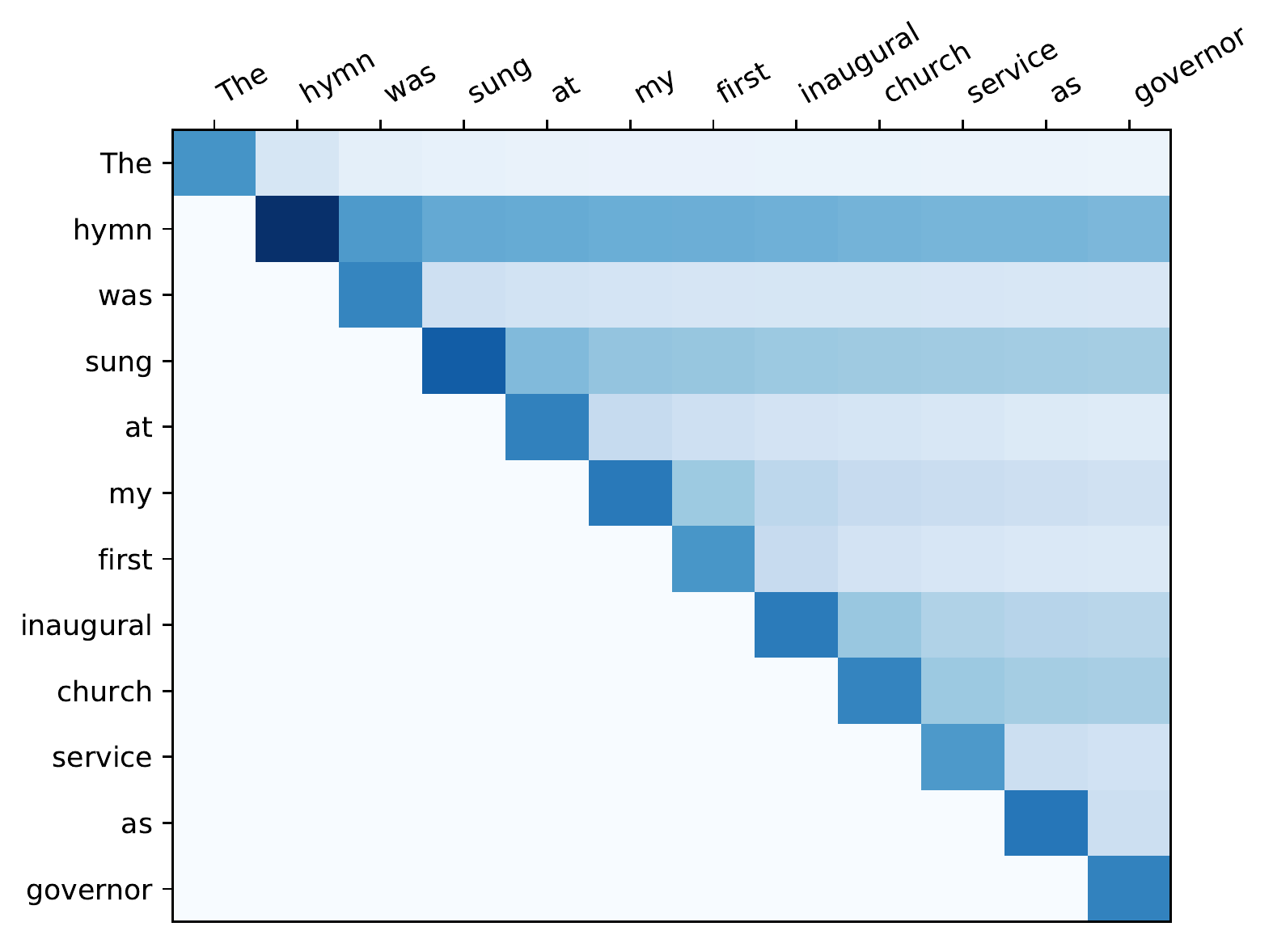} & \includegraphics[width=0.46\linewidth]{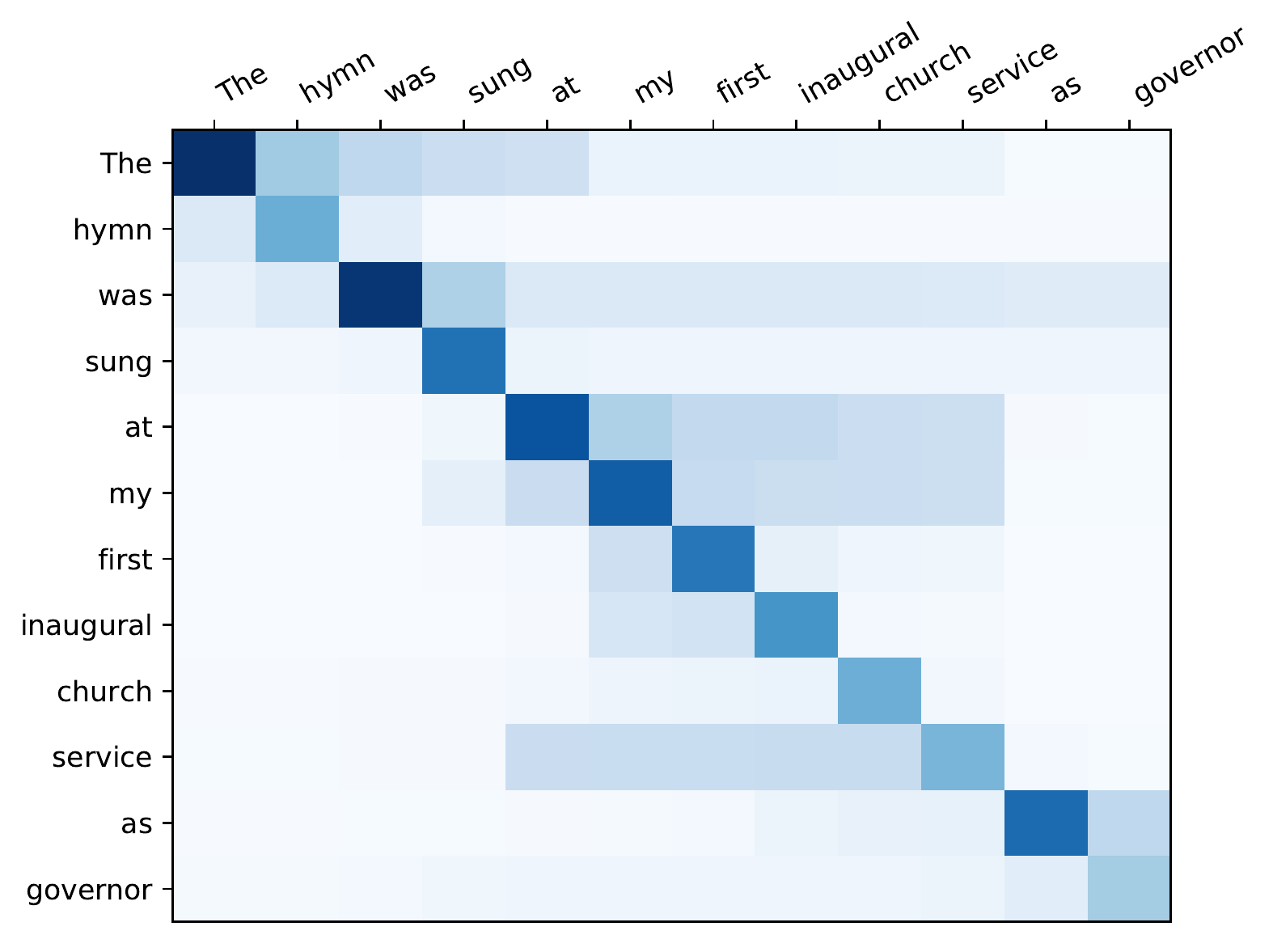} \\
    \hspace{-0.1cm}\includegraphics[width=0.46\linewidth]{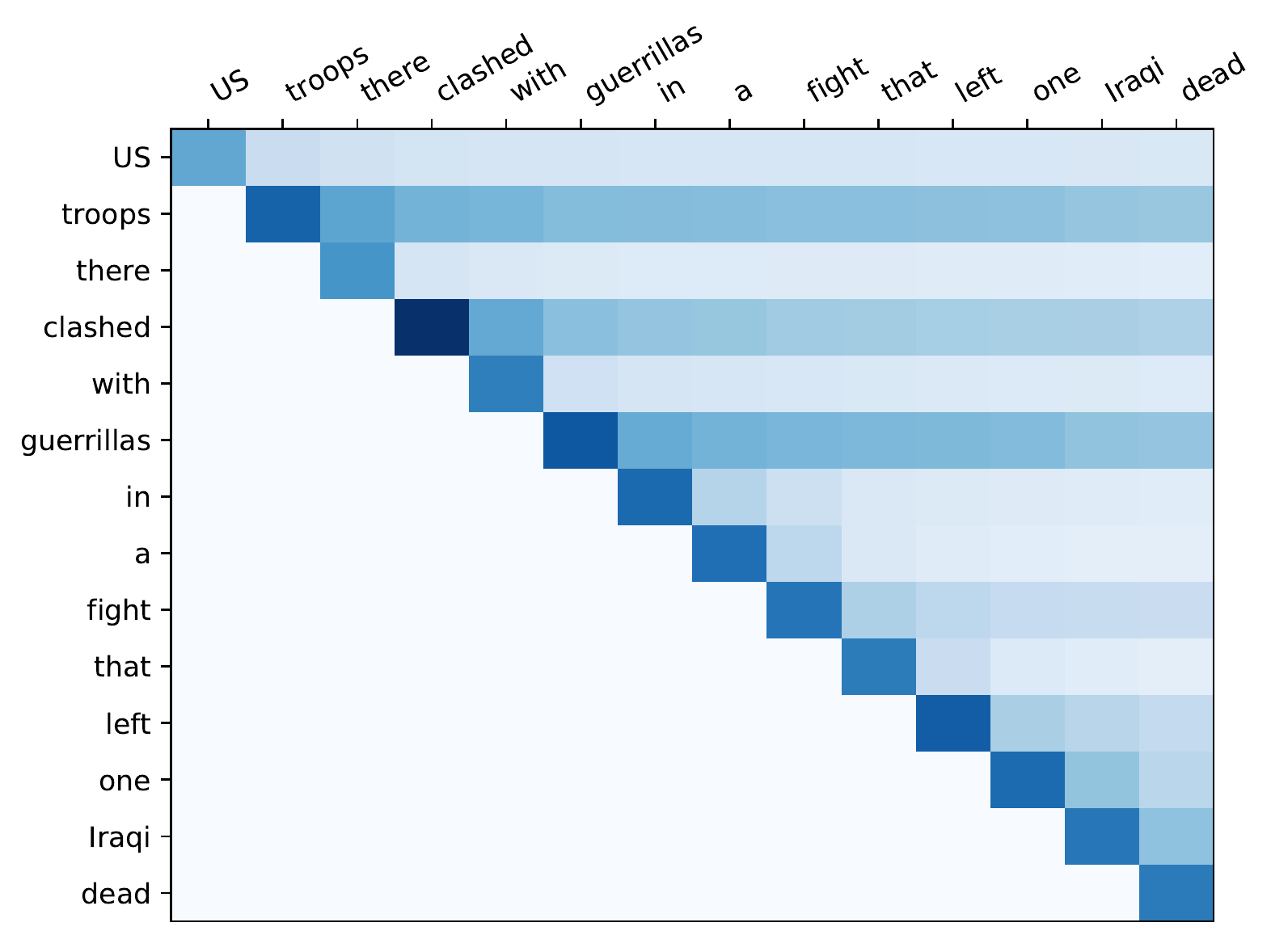} & \hspace{-0.1cm}\includegraphics[width=0.46\linewidth]{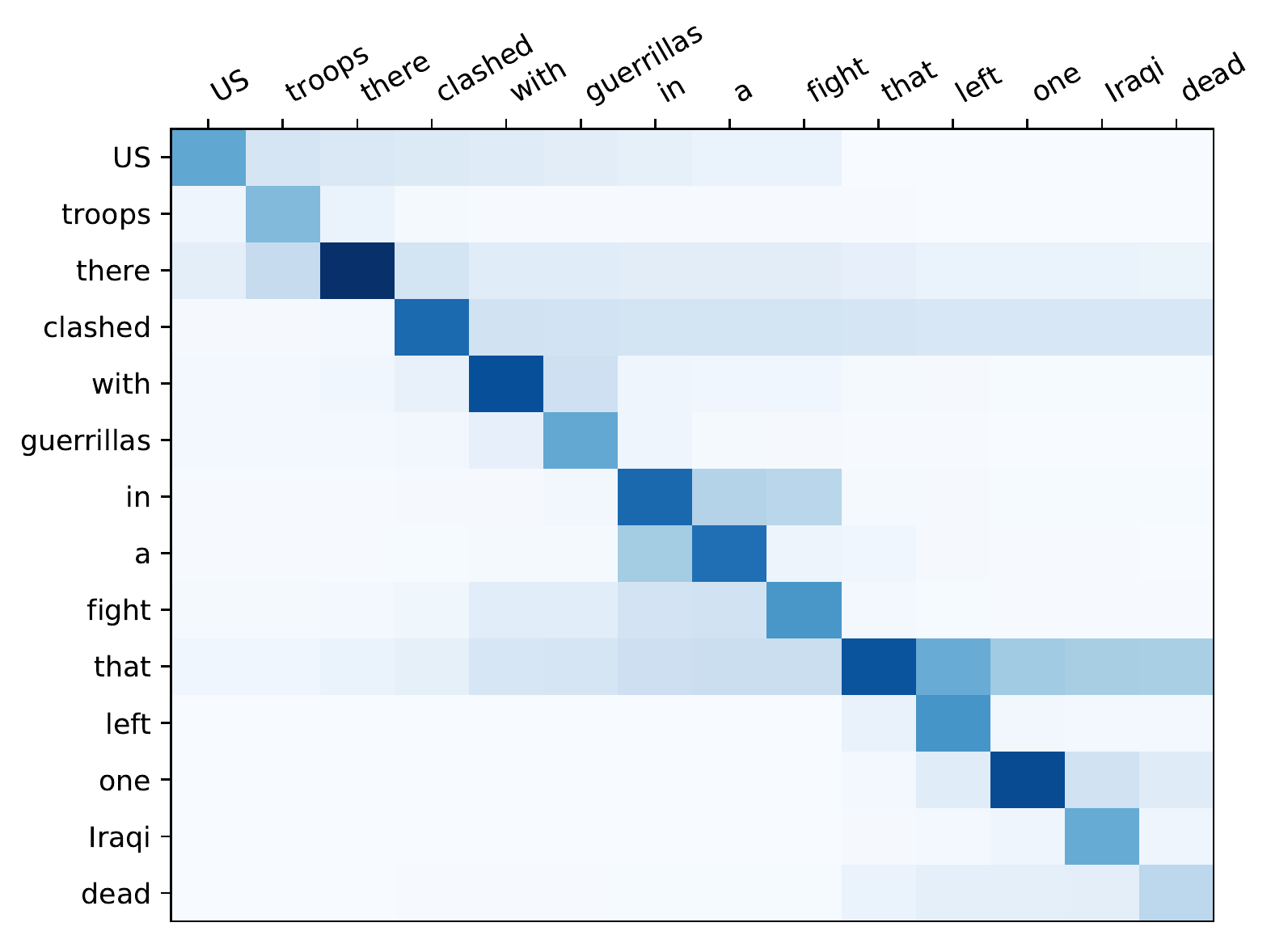} \\
    \hspace{-0.2cm}\includegraphics[width=0.46\linewidth]{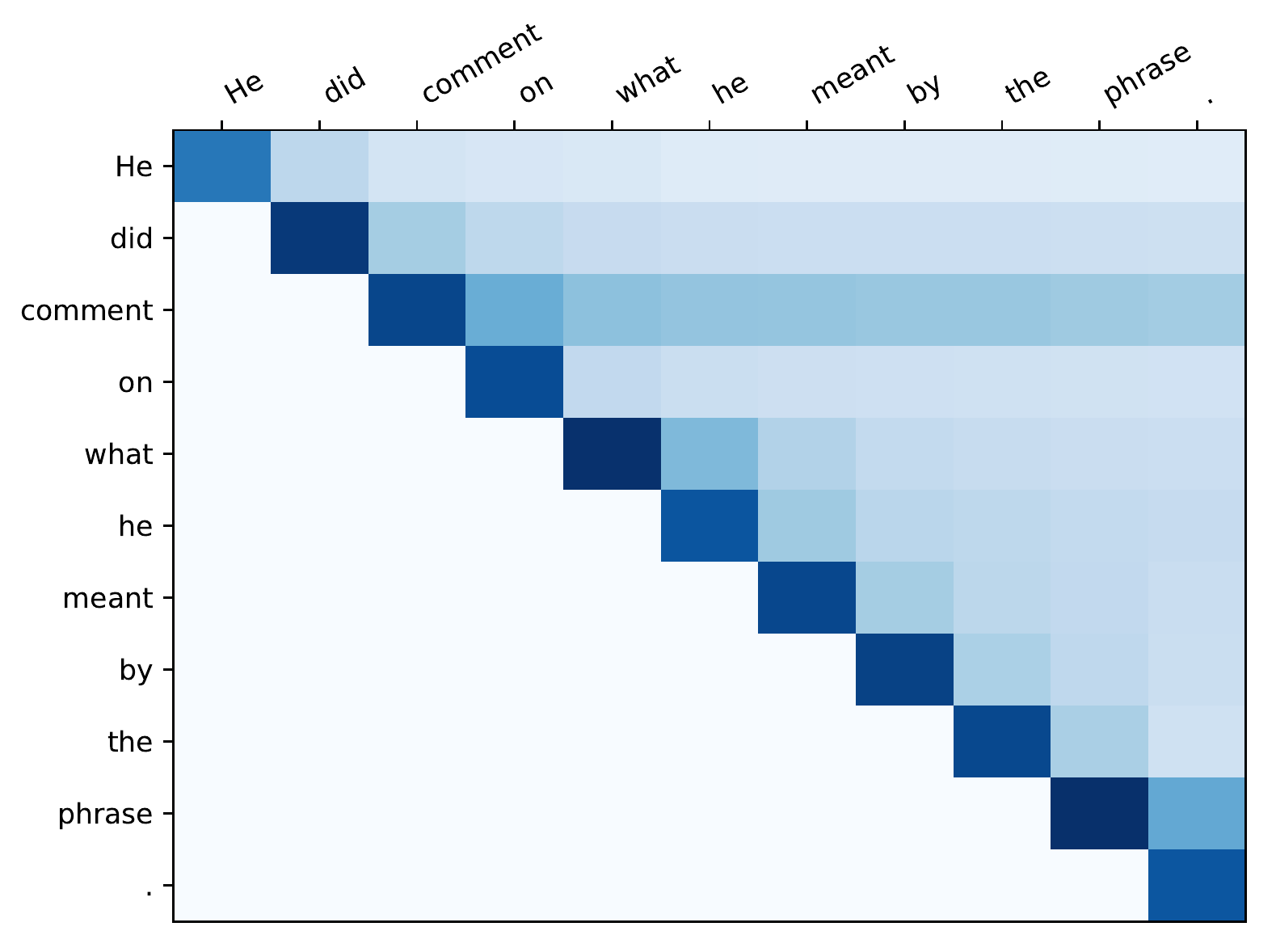} & \hspace{-0.2cm}\includegraphics[width=0.46\linewidth]{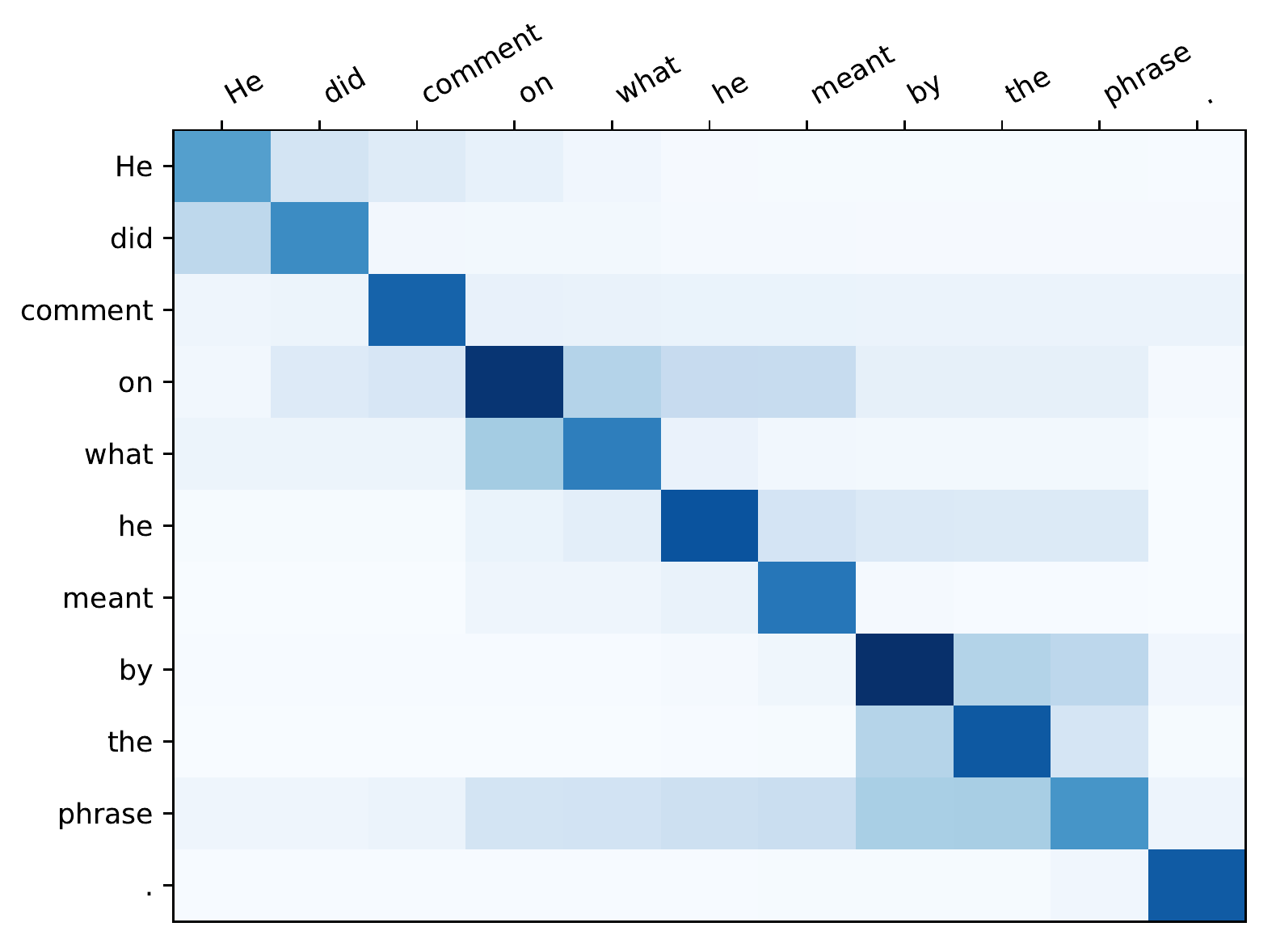} \\
    \hspace{-0.2cm}\includegraphics[width=0.46\linewidth]{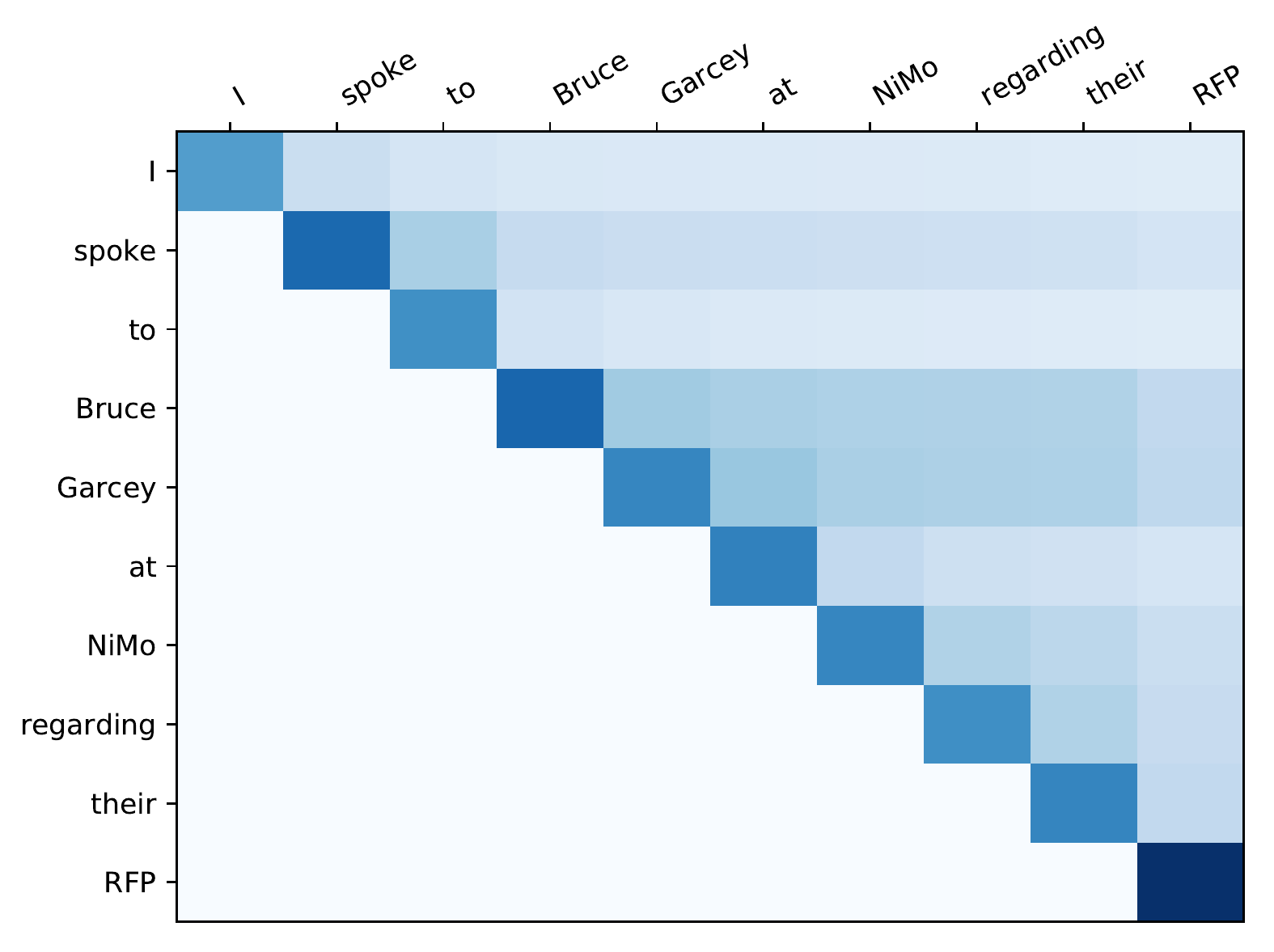} & \hspace{-0.2cm}\includegraphics[width=0.46\linewidth]{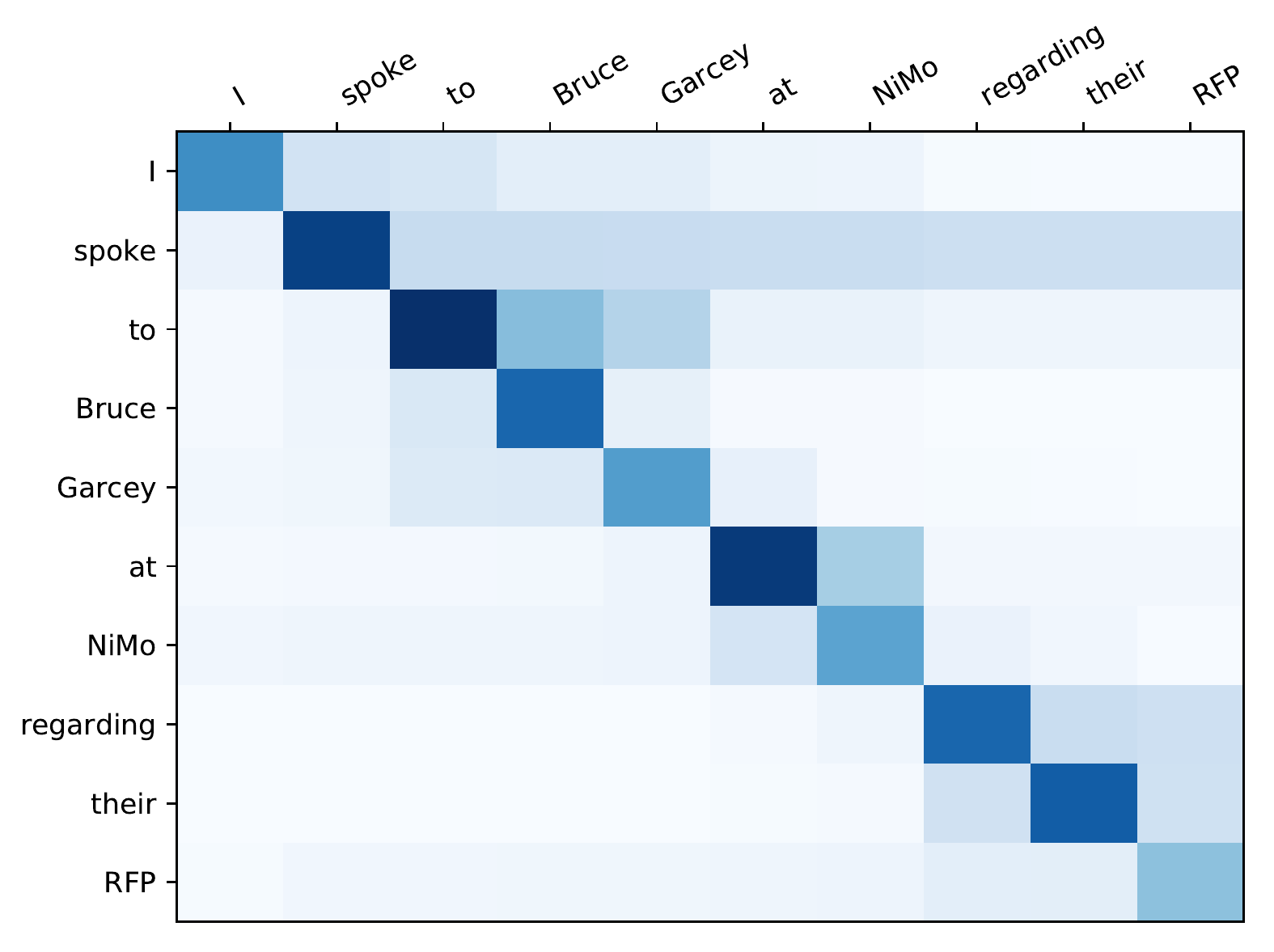} \\
        \bottomrule
    \end{tabular}
    \caption{Visualization of the weights on context words learned by the memory cell. Each column represents the current word $t$, and each row represents a context word $j$. The gating mechanism implicitly computes element-wise weighted sums over each column. The darkness of each square indicates the $L^2$-norm of the vector weights $\V{w}_{j}^{t}$ from Equation~ \ref{eq:weighted_sum_property}. Figures on the left show weights learned by a language model. Figures on the right show weights learned by a dependency parser.}
    \label{tab:viz}
\end{table*}

Given the empirical evidence that LSTMs are effectively learning weighted sums of the content layers, it is natural to investigate what weights the model learns in practice. Using the more mathematically transparent simplification of LSTMs, we can visualize the weights $\V{w}_{j}^{t}$ that are placed on every input $j$ at every timestep $t$ (see Equation~ \ref{eq:weighted_sum_property}).

Unlike attention mechanisms, these weights are vectors rather than scalar values. Therefore, we can only provide a coarse-grained visualization of the weights by rendering their $L^2$-norm, as shown in Table~\ref{tab:viz}. In the visualization, each column indicates the word represented by the weighted sum, and each row indicates the word over which the weighted sum is computed. Dark horizontal streaks indicate the duration for which a word was remembered. Unsurprisingly, the weights on the diagonal are always the largest since it indicates the weight of the current word. More interesting task-specific patterns emerge when inspecting the off-diagonals that represent the weight on the context words.

The first visualization uses the language model. Due to the language modeling setup, there are only non-zero weights on the current or previous words. We find that the common function words are quickly forgotten, while infrequent words that signal the topic are remembered over very long distances.

The second visualization uses the dependency parser. In this setting, since the recurrent architectures are bidirectional, there are non-zero weights on all words in the sentence. The top-right triangle indicates weights from the forward direction, and the bottom-left triangle indicates from the backward direction. For syntax, we see a significantly different pattern. Function words that are useful for determining syntax are more likely to be remembered. Weights on head words are also likely to persist until the end of a constituent.

This illustration provides only a glimpse into what the model is capturing, and perhaps future, more detailed visualizations that take the individual dimensions into account can provide further insight into what LSTMs are learning in practice.

\end{document}